\documentclass{ws-procs11x85}
\usepackage{ws-procs-thm}           

\newenvironment{myquote}{\list{}{\leftmargin=0.2in\rightmargin=0.2in}\item[]}{\endlist} 

\usepackage{multirow}

\usepackage{float}
\floatstyle{plaintop}
\restylefloat{table}

\usepackage{hyperref}
\hypersetup{
    colorlinks=true,
    linkcolor=blue,
    filecolor=blue,      
    urlcolor=blue,
}
\begin{document}

\title{Clinical Concept Embeddings Learned from Massive Sources of Multimodal Medical Data}
\begin{center}
\footnotesize{
    Andrew L. Beam\textsuperscript{1*\dag}}, Benjamin Kompa\textsuperscript{2*}, Allen Schmaltz\textsuperscript{1}, Inbar Fried\textsuperscript{3}, Griffin Weber\textsuperscript{2}, Nathan Palmer\textsuperscript{2}, Xu Shi\textsuperscript{1}, Tianxi Cai\textsuperscript{1}, Isaac S. Kohane\textsuperscript{2}\\ 
\small\itshape{$^1$Harvard T.H. Chan School of Public Health Boston, MA 02115, USA} \\ 
\small\textit{$^2$Harvard Medical School
Boston, MA 02115, USA} \\ 

\small\itshape{$^3$University of North Carolina School of Medicine\\
Chapel Hill, NC 27516, USA} \\ 
\textsuperscript{*}Denotes equal contribution
$^\dag$E-mail: andrew\_beam@hms.harvard.edu
\end{center}

\begin{abstract}
Word embeddings are a popular approach to unsupervised learning of word relationships that are widely used in natural language processing. In this article, we present a new set of embeddings for medical concepts learned using an extremely large collection of multimodal medical data. Leaning on recent theoretical insights, we demonstrate how an insurance claims database of 60 million members, a collection of 20 million clinical notes, and 1.7 million full text biomedical journal articles can be combined to embed concepts into a common space, resulting in the largest ever set of embeddings for 108,477 medical concepts. To evaluate our approach, we present a new benchmark methodology based on statistical power specifically designed to test embeddings of medical concepts. Our approach, called \emph{cui2vec},  attains state-of-the-art performance relative to previous methods in most instances. Finally, we provide a downloadable set of pre-trained embeddings for other researchers to use, as well as an online tool for interactive exploration of the \emph{cui2vec} embeddings. 
\end{abstract}


\copyrightinfo{\copyright\ 2019 The Authors. Open Access chapter published by World Scientific Publishing Company and distributed under the terms of the Creative Commons Attribution Non-Commercial (CC BY-NC) 4.0 License.}

\section{Introduction}
Word embeddings have become an extremely popular way to represent sparse, high-dimensional data in machine learning and natural language processing (NLP). Modern notions of word embeddings based on neural networks have their roots in the neural language model of Bengio et al. \cite{Bengio2003}, though the idea is closely related to many other approaches, notably latent semantic analysis (LSA) \cite{Berry1995} and hyperspace analogue to language (HAL) \cite{Lund1996}. Word embeddings are motivated by the observation that traditional representations for words, such as a one-hot encoding, are high dimensional and inefficient, since such an encoding captures none of the similarity or correlation information between words in the source text. The central idea is that a word can be characterized by ``the company it keeps,'' \cite{Harris1954} thus context words which appear around a given word encode a large amount of information regarding that word's meaning. Word embeddings model this contextual information by creating a lower-dimensional space such that words that appear in similar contexts will be nearby in this new space. 

The embedding approach in \emph{word2vec} \cite{Mikolov2013} has become quite popular since its introduction, and embeddings are now standard components in many NLP tasks. The main application has been in the use of ``transfer learning,'' where embeddings are first learned using extremely large sources of unlabeled text (from web-crawls, Wikipedia dumps, etc.), and the embeddings are then used in a supervised task as components of a model (e.g., a recurrent neural network) which accepts the pre-trained embeddings as inputs. It has been shown that transfer learning can work as well as it does for image data \cite{howard2018fine}, opening up numerous possibilities to exploit transfer learning in many NLP applications. Within the context of medical data, recent examples have shown that transfer learning works very well for imaging tasks \cite{Gulshan2016,Beam2016a}, due in large part to the availability of pre-trained computer vision models\cite{he2016deep,szegedy2016rethinking,simonyan2014very} that were pre-trained on the ImageNet database \cite{deng2009imagenet}. 

Machine learning has enormous potential in healthcare \cite{beam2018big}; however, many researchers lack access to large sources of non-imaging healthcare data due to privacy concerns. This has resulted in a lack of pre-trained resources for applications in healthcare and medicine relative to other areas of machine learning and NLP. Moreover, because healthcare data come in a variety of forms, popular word embedding algorithms like \emph{word2vec} and \emph{GloVe} \cite{Pennington2014}, which were originally developed for text, cannot be directly applied to many kinds of healthcare data. 

The primary goal of this work is to construct a comprehensive set of embeddings for medical concepts, which we refer to as \emph{cui2vec}, by combining extremely large sources of multimodal healthcare data.

\section{Overview of word2vec and GloVe}
\subsection{word2vec}
The original work that introduced \emph{word2vec} \cite{Mikolov2013} actually contains a collection of models and algorithms including the continuous bag of words (CBOW) model and the skip-gram model. The CBOW model predicts the probability of the target word given its context defined within a window, while the skip-gram model predicts the surrounding context given the target word. Specifically, the skip-gram model \cite{Mikolov2013} seeks to construct vector representations of a target word \emph{w} and a context word \emph{c} such that the conditional probability $p(w|c)$ is high for \textless$w,c$\textgreater~ pairs that co-occur frequently in the source text. For the remainder of this paper we will use \emph{w} and \emph{c} to refer to the target word and context word respectively, and use and $\vec{w}$, $\vec{c}$ to refer to the $1 \times d$ dimensional target word and context embeddings. Under the skip-gram model, the conditional probability of observing context word \emph{c} within a fixed window given the target word \emph{w} is proportional to the dot-product of their corresponding vectors, and is given by the \emph{softmax} function below:
\begin{align}
p(w|c) = \frac{\exp(\vec{w}\vec{c}^T)}{\sum_j\exp(\vec{w}\vec{c_j}^T)}
\end{align}

where the sum in the denominator is over all unique context words in the source corpus. Note that this sum is generally intractable and requires approximations to estimate efficiently. Thus, the vectors $\vec{w}, \vec{c}$ encode information about how likely word \emph{w} is to appear in a randomly selected piece of text, given word \emph{c} has been observed.

A key feature of \emph{word2vec} are techniques that enable efficient training on large corpora. For example, negative sampling approximates the sum in the denominator by randomly sampling \emph{k} context words which do not appear in the current window. This allows the algorithm to be run with bounded memory requirements and in a parallel fashion, which improves the training speed and enables training on very large corpora \cite{mikolov2013efficient}. Indeed, the key point of Mikolov et al. was that training a simple and scalable model with more data results in better accuracy than a complex non-linear model on a variety of benchmarks.

\subsection{GloVe}
Global Vectors for Word Representations (GloVe) \cite{Pennington2014} was introduced shortly after Mikolov et al. and differs in several important ways. \emph{GloVe} produces word embeddings by fitting a weighted log-linear model to co-occurrence statistics. Given that a target word \emph{w} and a context word \emph{c} co-occur $y$ times, \emph{GloVe} solves the following least-squares optimization problem:
\begin{align}
\underset{\vec{w},\vec{c}, b_w, b_c}{\mathrm{argmin}} f(y)(\vec{w}\vec{c}^T + b_w + b_c - \log(y))^2
\end{align}
where $b_w,b_c$ are word and context biases, respectively and $f(y)$ is a weighting function and is given by:
\begin{align}
f(y) = \left\{ \begin{array}{cc} 
                \left(\frac{y}{y_{max}}\right)^\alpha & \hspace{5mm} y<y_{max} \\
                1 & \hspace{5mm} y \geq y_{max} 
                \end{array} \right.
\end{align}
The final embedding for word $i$ is the sum of the resulting word and context vectors for that word. This is repeated for all \emph{w},\emph{c} pairs and is trained iteratively using stochastic gradient descent. The most expensive step is the construction of the term-term co-occurrence matrix, which is necessary before training can begin.

\subsection{Embeddings as a Factorization of a Modified Co-occurrence Matrix}
Previous work \cite{Levy2014} by Levy and Goldberg showed that the skip-gram model with negative sampling (SGNS), which is often considered to be state-of-the-art \cite{Baroni2014}, is implicitly factorizing a shifted, positive pointwise mutual information (PMI) matrix of word-context pairs. Pointwise mutual information (PMI) is a measure of association between a word and a context word, and can be computed from the counts of word-context pairs in the corpus, given by:
\begin{align} \label{pmi}
\textrm{PMI}(w,c) = \frac{p(w,c)}{p(w)*p(c)}
\end{align}
where $p(w,c)$ is the number of times word \emph{w} and context-word \emph{c} occur in the same context window divided by the total number of word-context pairs, whereas $p(w)$, $p(c)$ are the singleton frequencies of \emph{w} and \emph{c}, respectively. If we shift the PMI by some constant $\log(k)$ (where $k$ is the number of negative samples in the original \emph{word2vec} paper\cite{Mikolov2013}) and set all negative entries to 0, and factor the resulting \emph{shifted positive pointwise mutual information matrix} (SPPMI) we recover the implicit objective of \emph{word2vec}'s SGNS model. The element wise SSPMI transformation is shown below:
\begin{align}
\textrm{SPPMI}(w,c) = \max(\textrm{PMI}(w,c) - \log(k),0)
\end{align}
Therefore, one can simply factorize the SSPMI matrix using any factorization method, such as a singular value decomposition (SVD), to obtain a lower-dimension embedding of the words. This finding is critical as it links \emph{word2vec} to traditional count-based methods that are based on co-occurrence statistics. 

\emph{GloVe} was originally presented in terms of explicit matrix factorization and provides an algorithm to perform this factorization (stochastic gradient descent to minimize sum-of-squared error). Thus, under this unified framework the starting point for both \emph{word2vec} and \emph{GloVe} is the construction of a term-term co-occurrence matrix. This insight is what allows us to use these algorithms on problems which may contain non-textual data sources, as we can materialize a co-occurrence matrix using any data where such co-occurrences can be computed. Then we simply use the \emph{GloVe} algorithm to directly factor this matrix or use SVD to factor the SSMPI matrix to create \emph{word2vec} style embeddings.

\subsection{Overview of cui2vec}
Medical data are multi-modal by nature and come in many forms including free text (in medical publications and clinical notes) and billing codes for diagnoses and procedures in the electronic healthcare record (EHR). The \emph{cui2vec} system works by first mapping all of these concepts into a common concept unique identifier space (CUI) using a thesaurus from the Unified Medical Language System (UMLS). Next, a CUI-CUI co-occurrence matrix is constructed, but the way a co-occurrence is counted depends on the source data. For non-clinical text data (e.g., journal articles), it is first preprocessed (see Section \ref{Sec:Methods}) and chunked into fixed length windows of 10 words, and a co-occurrence is counted as the appearance of a CUI-CUI pair in the same window. For claims data, ICD-9 codes are mapped to UMLS CUIs and a co-occurrence is counted as the number of patients in which two CUIs appear in any 30-day period.  Finally, for the clinical notes, we counted a co-occurrence as two CUIs appearing in the same 30-day `bin' in a similar fashion to previous work\cite{Choi2016}, but see the original publication \cite{Finlayson2014} for the precise definition. Once the master co-occurence matrix is created, it can be directly factored by \emph{GloVe} or transformed into a SSPMI matrix and factored using SVD to create \emph{word2vec} embeddings. 

\subsubsection*{Related Work}
There is a long history of machine learning and natural language processing for clinical uses, but for the purposes of this paper we confine our review to papers that are directly seeking to create low dimensional representations of clinical concepts, in the spirit of \emph{word2vec} and \emph{GloVe}. The first investigations \cite{Minarro-Gimenez2014,DeVine2014,moen2013distributional} using \emph{word2vec} for medical concepts were performed shortly after the original \emph{word2vec} paper appeared in 2013 and reported mixed results, though De Vine et al. reported state-of-the-art performance with respect to human assessments of concept similarity and relatedness. 

Liu et al. \cite{liu2015exploiting} used embeddings jointly trained on Wikipedia and ICU notes to perform automatic expansion of abbreviations which are common in clinical notes. Lastly, Choi et al. \cite{Choi2016} performed the work that is most comparable to this study, which used similar sources of data to create embeddings for UMLS CUIs. Choi et al. used a claims database of 4 million patients and a novel methodology to create a set of clinical embeddings as well as the notes from Finlayson et al. \cite{Finlayson2014}

\subsection{Contributions of this work}
The work presented here differs in several important ways from existing works. First, we have access to a much larger claims database of 60 million patients and a larger set of 1.7 million full text articles (not restricted to abstracts), which should enable both a much larger and higher quality set of embeddings. Secondly, the embeddings produced by Choi et al. are different for each data source, whereas we map all concepts into a common co-occurrence space to produce a single set of embeddings that can be used on tasks with different kinds of clinical data.  We also present a new and expanded evaluation methodology that is both more interpretable and, we believe, a more natural way to benchmark sets of clinical embeddings that will be of general use for future medical embedding work. Finally, we believe that our approach incorporates many of the best practices with respect to tuning parameters (see Section~\ref{Sec:Methods}) which also results in increased performance. In summary, this work presents results in a new set of embeddings for 108,477 medical concepts, the largest ever such collection, which are derived from three sources of clinical data and are equal to or exceed the existing state of the art on nearly all benchmarks.

\section{Materials and Methods}\label{Sec:Methods}
\subsection{Data Sources}
The data come from the following three independent sources: an un-identifiable claims database from a nationwide US health insurance plan with 60 million members over the period of 2008-2015, a dataset of concept co-occurrences from 20 million notes at Stanford \cite{Finlayson2014}, and an open access collection of 1.7 million full text journal articles obtained from PubMed Central. For the purposes of this study, the insurer has asked not to be named. 

\subsection{Text Normalization and Preprocessing}
For text data it is important to first normalize against some standard vocabulary or thesaurus. Word embeddings operate on tokens, and many medical concepts can span multiple tokens. To collapse multi-word concepts into a single token, we used the Narrative Information Linear Extraction (NILE) \cite{yu2013short} system normalized against the Systematized Nomenclature of Medicine - Clinical Terms (SNOMED-CT) \cite{donnelly2006snomed} reference thesaurus. SNOMED-CT IDs were then mapped to concept unique identifiers (CUIs) from the UMLS \cite{bodenreider2004unified}. The pipeline converts all letters to lowercase, removes punctuation, and replaces all medical concepts with their CUI representation (e.g., `bronchopulmonary dysplasia' with C0006287 and `resulting from' with C0678226). For example, our pipeline would transform the following sentence (taken from previous work\cite{Jobe2001}):
\begin{small}
\begin{myquote}
\texttt{Bronchopulmonary Dysplasia was first described by Northway and colleagues in 1967 as a lung injury in a preterm infant resulting from oxygen and mechanical ventilation.}
\end{myquote}
\end{small}
into the following normalized representation: 
\begin{small}
\begin{myquote}
\texttt{C0006287 was first described by northway and colleagues in 1967 as a C0024109 C3263722 in a C0021294 C0678226 C0030054 and C0199470}
\end{myquote}
\end{small}
\subsubsection*{Benchmarks and Evaluation}
The benchmarking strategy leverages previously published `known' relationships between medical concepts. We compare how similar the embeddings for a pair of concepts are by computing the cosine similarity of their corresponding vectors, and use this similarity to assess whether or not the two concepts are related. Cosine similarity between word vectors $\vec{w_1}, \vec{w_2}$ is given by: 
\begin{align*}
\cos(\vec{w_1},\vec{w_2}) = \frac{w_1w^T_2}{\|w_1\|_2\|w_2\|_2}
\end{align*}
and is 1 if the vectors are identical and 0 if they are orthogonal. One approach would be to rank the cosine similarity for a known relationship against all others via a ranking metric such as mean-precision or discounted cumulative gain. However, such a strategy has several limitations. The primary issue is that many concepts may correctly be ranked higher than the query concept, but they may not be part of the database of known relationships. Thus, a ranking metric may incorrectly penalize a set of embeddings simply because some true relationships were ranked higher but were not included in the list of `known' relationships. 

Instead, we present a new approach based on the notion of statistical power. For a known relationship pair $(x,y)$, we first compute the null distribution of scores by drawing 10,000 bootstrap samples $(x^*,y^*)$ where $x^*$ and $y^*$ belong to the same category as $x$ and $y$, respectively. For example, when assessing whether `preterm infant' (which is a disease or syndrome) is associated with `bronchopulmonary dysplasia' (also a disease or syndrome), we would randomly sample two concepts from the ``disease or syndrom'' class and compute their cosine similarity, and then repeat this procedure 10,000 times to create the bootstrap null distribution. We then compare the observed score between x and y and declare it statistically significant if it is greater than the 95th percentile of the bootstrap distribution (e.g., $p < 0.05$ for a one-sided test). Applying this procedure to the collection of known relationships, we calculate the statistical power to reject the null of no relationship which is the quantity we report in all experiments, except for the comparison to human assessments of similarity. This metric has the added benefit of being easy to interpret, as it is an estimate of the fraction true relationships discovered given a tolerance for a 5\% false positive rate. 

Below is a list of the benchmarks used in this study, along with details that are specific to each. We provide an example of a known relationship from each category to help the reader understand the types of relationships captured by each benchmark.
\begin{itemize}
  \item \textbf{Comorbid Conditions}: A comorbidity is a disease or condition that frequently accompanies a primary diagnosis. We created a curated set of comorbid conditions for Addison's disease, autism, heart disease, obesity, schizophrenia, type 1 diabetes and type 2 diabetes. These comorbidities were extracted from the Mayo Clinic's Encyclopedia of Diseases and Conditions \cite{Staff}, Wikipedia, and the Merck Manuals \cite{beers1999merck}. 
  \begin{itemize}
     \item \emph{Example}: Primary condition: premature infant (CUI: C0021294) Comorbidity: bronchopulmonary dysplasia (CUI: C0006287)
   \end{itemize}
  \item \textbf{Causative Relationships}: The UMLS contains a table (MRREL) of entities known to be the cause of a certain result. We extracted known instances of the relationships \emph{cause of} and \emph{causative agent}, and \emph{induces} from the MRREL table. We computed the null distribution for these relationships by computing the similarity of randomly sampled concepts with the same semantic type as the cause and randomly sampled concepts with the same semantic type as the result. 
    \begin{itemize}
       \item \emph{Example}: Cause: Jellyfish sting (CUI: C0241955) Result: Irukandji syndrome (CUI: C1655386)
     \end{itemize}
	\item \textbf{National Drug File Reference Terminology (NDF-RT)}: The NDF-RT was created by the U.S. Department of Veterans Affairs, Veterans Health Administration. We extracted drug-condition relationships using the \emph{may prevent} and \emph{may treat} relationships. We assessed power to detect \emph{may treat} and \emph{may prevent} relationships using bootstrap scores of random drug-disease pairs. 
  \begin{itemize}
     \item \emph{Example}: Drug: abciximab (CUI: C0288672) May Treat: Myocardial Ischemia (CUI: C0151744)
   \end{itemize}
	\item \textbf{UMLS Semantic Type}: Semantic types are meta-information about which category a concept belongs to, and these categories are arranged in a hierarchy. We extracted the most specific semantic type available for each concept from the MRSTY file provided by UMLS. To assess power to detect if two concepts belonged to the same semantic type, we randomly sampled concepts from different semantic type classes and computed a marginal null distribution of scores.
  \begin{itemize}
     \item \emph{Example}: Concept: Metronidazole (CUI: C0025872, Semantic Type: Pharmacologic Substance) Concept: Clofazimine (CUI: C0008996, Semantic Type: Pharmacologic Substance)
   \end{itemize}
  	\item \textbf{Human Assessment of Concept Similarity}: Previous work \cite{pakhomov2010semantic} has assessed how resident physicians perceive relationships among 566 pairs of UMLS concepts. Each concept pair has an average measure of how similar or related two concepts are to be as judged by resident physicians. We report Spearman correlation between human assessment scores and cosine similarity from the embeddings for this benchmark. 
\end{itemize}
\subsection{Implementation Details}
There are many hyper-parameters associated with both \emph{word2vec} and \emph{GloVe} that can have a dramatic effect on performance. In \emph{word2vec} parameters such as the number of negative samples, the size of the context window, the amount of smoothing for the context singleton-frequencies, and whether or not the context vectors are used to construct the final embeddings are all options that the practitioner must choose. Levy and Goldberg \cite{Levy2015} conducted a systematic set of experiments on the effects of these hyper-parameters on the performance of \emph{word2vec}, and we follow their recommendations in this work. Specifically, we used the following settings for all \emph{word2vec} experiments that are based on a singular value decomposition (SVD):
\begin{itemize}
\item	Smoothing of singleton frequencies by a constant exponential term. Instead of using $p(w)$ in (\ref{pmi}), we instead use $p(w)^\alpha$, where $\alpha$ is set to 0.75. In Levy and Goldberg, they recommend only smoothing the context singleton frequencies, but our co-occurrence matrices are symmetric so there is no difference in the singleton frequency when it is a `word' and when it is a `context'.
\item We set $k=1$ in the SPPMI transformation (i.e., no shift). 
\item We construct the final embeddings using a symmetrically scaled sum of the word and context vectors resulting from the singular value decomposition. Given the first $d$ singular vectors and singular values resulting from the SVD of a SPPMI matrix $X$, $SVD_d(X) = U_d\Sigma_dV_d$, the $d$-dimensional word embeddings $W$ are constructed as follows:
\begin{align*}
\tilde{W} &= U_d\sqrt{\Sigma_d}\\
\tilde{C} &= V_d\sqrt{\Sigma_d}\\
W &= \tilde{W} + \tilde{C}
\end{align*}
\item The SVD of the sparse SPPMI matrix was performed using the \emph{augmented implicitly restarted Lanczos bidiagonalization algorithm} with the \texttt{irlba} package \cite{Baglama2005,Baglama2017} in the R programming language.
\end{itemize}
For the comparison to the traditional \emph{word2vec} algorithm on the articles from PubMed, we used the implementation available in the 
\texttt{gensim} python package \cite{rehurek2010software}. We used the skip-gram algorithm, hierarchical softmax, 10 negative samples, and a window size of 10. We used the implementation of GloVe available in the R package \texttt{text2vec} \cite{selivanov2016text2vec}. We used the sum of target word and context vectors as the final embedding and set the $y_{max}=100$. As a baseline, we performed a SVD on the raw co-occurence matrix, and we report these results as \emph{PCA}. 

\section{Results}
\subsection{Benchmark Results}
We compared embeddings created by \emph{GloVe}, \emph{word2vec}, and \emph{PCA} on our benchmarks to determine which algorithm and embedding dimension produced the best results across each individual dataset and on the combined data. These results are shown in Table 1. The best configuration was \emph{word2vec} with an embedding dimension of 500, as it achieved the highest performance across nearly all benchmarks. Interestingly, we saw only a modest effect of embedding dimension on the benchmarks based on power (see Supplement). Also of note, the most direct comparison we could make to the original \emph{word2vec} algorithm was using PubMed articles. On this dataset, \emph{word2vec} based on a SVD was better than the original algorithm, as shown in the second row group in Table \ref{table:benchmarktable}. 


\begin{table}[ht]
\centering
\scalebox{0.8}{
\begin{tabular}{llcccc|c}
  \hline
Data Source & Algorithm & Causative & Comorbidity & Semantic Type & NDFRT & Human Assessment  \\ 
  \hline
\multirow{3}{*}{Claims} & GloVe & \textbf{0.56} & \textbf{0.73} & 0.29 & - & \textbf{0.45}  \\ 
  & PCA & 0.40 & 0.15 & 0.32 & - & 0.19 \\ 
  & word2vec (SVD) & 0.54 & 0.50 & \textbf{0.40} & - & \textbf{0.45} \\ 
   \hline
  \multirow{4}{*}{PMC Articles} &  GloVe & 0.59 & 0.57 & 0.28 & 0.54 & 0.60 \\ 
  & PCA & 0.30 & 0.24 & 0.24 & 0.29 & 0.29\\ 
  &  word2vec (SVD) & \textbf{0.83} & \textbf{0.59} & \textbf{0.49} & \textbf{0.84} & \textbf{0.67} \\ 
  & word2vec (original) & 0.75 & 0.51 & 0.48 & 0.74 & 0.59 \\ 
   \hline
  \multirow{3}{*}{Clinical Notes} & GloVe & 0.39 & \textbf{0.73} & 0.51 & 0.11 & 0.34 \\ 
  & PCA & 0.36 & 0.31 & 0.47 & 0.14 & 0.53 \\ 
  & word2vec (SVD) & \textbf{0.75} & 0.52 & \textbf{0.74} & \textbf{0.49} & \textbf{0.59} \\ 
   \hline
  \multirow{3}{*}{Combined Data} & GloVe & 0.40 & \textbf{0.80} & 0.37 & 0.50 & 0.39 \\ 
  & PCA & 0.24 & 0.23 & 0.30 & 0.37 & 0.47 \\ 
  & word2vec (SVD)  & \textbf{0.46} & 0.52 & \textbf{0.53} & \textbf{0.57} & \textbf{0.47} \\ 
   \hline
   
\end{tabular}
}
\caption{\small{Comparison of \emph{GloVe}, \emph{PCA}, and \emph{word2vec} for an embedding dimension of 500. Columns 1-4 report power to detect known relationships and column 5 reports the Spearman correlation between human assessments of concept similarity and cosine similarity from the embeddings. The best result for each each benchmark/dataset combination is shown in bold. The claims dataset contained only diagnosis codes and no drugs and so did not report results for the NDFRT benchmark.}\vspace{\baselineskip}}
\label{table:benchmarktable}
\end{table}

The 500-dimensional \emph{word2vec} style embeddings using the combined data are referred to as the \emph{cui2vec} embeddings in all subsequent experiments. 

\subsection{Comparison to previous results}
\begin{figure}[h]
  \centering
  \includegraphics[scale=0.35]{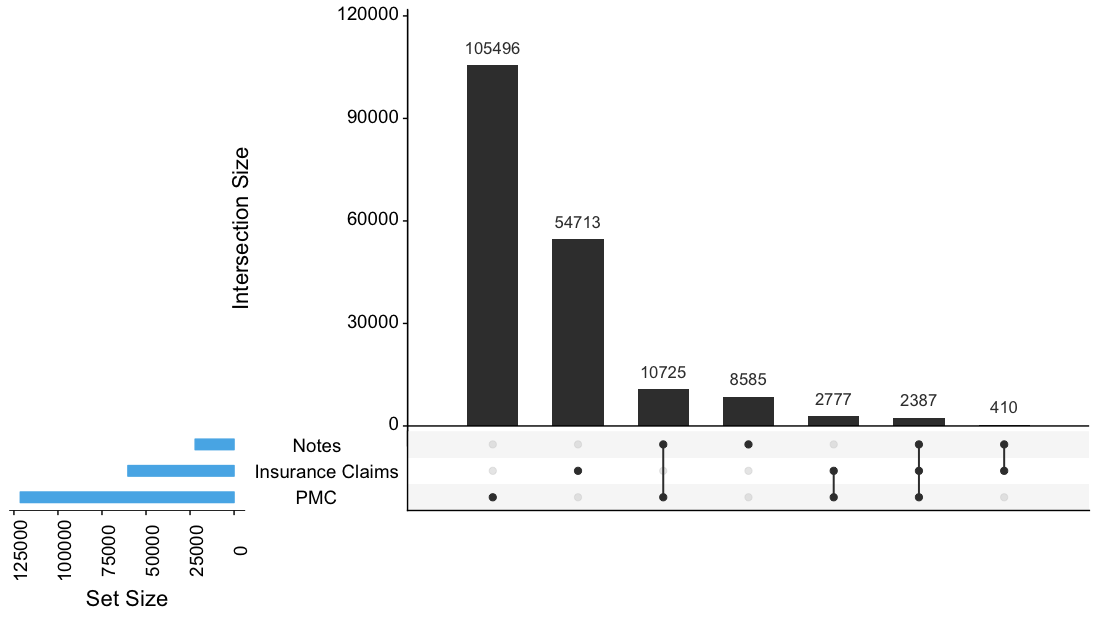}
  \caption{\emph{Upset} visualization of the intersection of medical concepts found in the insurance claims, clinical notes, and biomedical journal articles (PMC).}
  \label{upset}
\end{figure}

In total we were able to estimate embeddings for 108,477 unique concepts using the combined set of data, making this the largest set of embeddings for medical concepts to date. Figure \ref{upset} shows a visualization of the various intersections of the 108,477 concepts found across the different sources of data using the UpSet visualization method \cite{Conway2017,lex2014upset}. 

Most of the concepts appear in only one corpus, however 16,299 (14\%) appeared in multiple sources.  We evaluated previously published embeddings obtained through the \emph{clinicalml} github repository (\url{https://github.com/clinicalml/embeddings}) for comparison to our \emph{cui2vec} embeddings. Note that all three of the comparison embeddings come from different data sources and have very few concepts in common, so we were forced to perform pairwise comparisons between \emph{cui2vec} and each set of embeddings.

The first comparison was against 300-dimensional embeddings for 15,905 concepts (of which 12,568 were in common with \emph{cui2vec}) derived from a claims database of 4 million patients. The results are shown in Table \ref{comparetable}. We observed that \emph{cui2vec} outperformed the reference embeddings in most tasks, in some instances by a substantial margin, though the embeddings from Choi et al. had the edge in the human assessment benchmark. Next, we compared 300-dimension embeddings for 28,394 concepts derived from the same set of clinical notes in Finlayson et al.\cite{Finlayson2014} published as part of Choi et al.\cite{Choi2016} In total, there were 21,789 concepts in common between \emph{cui2vec} and this set of embeddings. Here \emph{cui2vec} was again better in most benchmarks, in some cases by a large margin.  Finally, we compared \emph{cui2vec} against 200-dimensional embeddings for 59,266 concepts derived from 348,566 PubMed abstracts, first published in De Vine et al.\cite{de2014medical} There were 33,376 concepts in common that were used for benchmarking. On this dataset we observed a huge relative improvement and \emph{cui2vec} was uniformly better across all benchmarks, as shown in Table \ref{comparetable}.

\begin{table}[ht]
\centering
\scalebox{0.8}{
\begin{tabular}{lcccc|c}
\hline
 Source & Causative & Comorbidity & NDFRT & Semantic Type & Human Assessment \\ 
 \hline
 Choi et al. (claims) & 0.25 & \textbf{0.37} & 0.63 & 0.24 & \textbf{0.47} \\ 
  cui2vec & \textbf{0.55} & 0.31 & \textbf{0.73} & \textbf{0.43} & 0.35 \\ 
   \hline
Choi et al. (notes) & 0.29 & 0.23 & \textbf{0.52} & 0.15 & 0.43 \\ 
  cui2vec & \textbf{0.42} & \textbf{0.25} & 0.42 & \textbf{0.36} & \textbf{0.51} \\ 
   \hline
Devine et al. (PMC abstracts) & 0.29 & 0.05 & 0.18 & 0.22 & 0.45 \\ 
  cui2vec & \textbf{0.48} & \textbf{0.31} & \textbf{0.46} & \textbf{0.48} & \textbf{0.50} \\ 
   \hline
\end{tabular}
}
\caption{\small{Comparison of the performance of \emph{cui2vec} to previously published embeddings. Columns 1-4 report power to detect known relationships and column 5 reports the Spearman correlation between human assessments of concept similarity and cosine similarity from the embeddings. The best result for each each comparison is shown in bold.}\vspace{\baselineskip}}
\label{comparetable}
\end{table}

\vspace*{-\baselineskip}

\subsection{Discussion}
In this study we have created the most comprehensive set of 108,299 clinical embeddings to date using extremely large and multi-modal sources of medical data. When compared to previous results, the \emph{cui2vec} embeddings achieve state-of-the-art performance in many instances. Even though there is more healthcare data than ever, most of it is either unlabeled or weakly labeled, so the ability to extract meaningful structure in an unsupervised manner is extremely important. Another potential obstacle is that most sources of healthcare data are not easily shareable, which limits some researchers to small sources of local data. We hope to reduce both of these barriers by providing our \emph{cui2vec} embeddings that were created using large and national sources of healthcare data. We believe that these embeddings will be generally useful for a variety of clinically oriented machine learning tasks.

\subsubsection*{Availability of Code and Data}
An R package \texttt{cui2vec} implementing the \emph{cui2vec} system can be found at the \href{https://github.com/beamandrew/cui2vec}{github repository}. An interactive explorer of the embeddings can be found \href{http://cui2vec.dbmi.hms.harvard.edu/}{here}.

\section*{Acknowledgements}
ALB was supported by NIH/NHLBI 7K01HL141771-02,  
BK was supported by NIH T32HG002295. 


\bibliographystyle{ws-procs11x85}
\bibliography{ws-pro-sample}

\end{document}